\documentclass[]{interact}

\usepackage{epstopdf}% To incorporate .eps illustrations using PDFLaTeX, etc.
\usepackage[caption=false]{subfig}% Support for small, `sub' figures and tables
\usepackage{lscape}

\usepackage[numbers,sort&compress]{natbib}% Citation support using natbib.sty
\bibpunct[, ]{[}{]}{,}{n}{,}{,}% Citation support using natbib.sty
\renewcommand\bibfont{\fontsize{10}{12}\selectfont}% Bibliography support using natbib.sty
\makeatletter% @ becomes a letter
\def\NAT@def@citea{\def\@citea{\NAT@separator}}% Suppress spaces between citations using natbib.sty
\makeatother% @ becomes a symbol again

\theoremstyle{plain}% Theorem-like structures provided by amsthm.sty

\theoremstyle{definition}

\theoremstyle{remark}

\begin{document}

\articletype{SURVEY PAPER}% Specify the article type or omit as appropriate

\title{Survey and Perspective on Social Emotions in Robotics}

\author{
\name{Chie Hieida\textsuperscript{a}\thanks{CONTACT Chie Hieida Email: hieida@is.naist.jp} and Takayuki Nagai\textsuperscript{b,c}}
\affil{\textsuperscript{a}Division of Information Science, Graduate School of Science and Technology, Nara Institute of Science and Technology, Japan; \textsuperscript{b}Department of Systems Innovation, Graduate School of Engineering Science, Osaka University, Japan; \textsuperscript{c}The University of Electro-Communications, Japan}
}

\maketitle

\begin{abstract}
%本論文は社会的感情とロボットに関して調査し、まとめたものである。ロボットにおいて、感情の研究は様々なアプローチで行われてきた。例えば、感情認識や感情表出、感情モデル研究などだ。これらの研究ではカテゴリ説や次元説などの心理学的知見に従い、研究が推進されてきた。しかしながら、これらの説を基にした研究の多くは基本的な感情のみしか扱えておらず、高次感情ともいわれる社会的感情は扱えていない。そこで、心理学や神経科学などの社会的感情の知見をまとめつつ、現状ロボットにおいて社会的感情がどのように実現されているのかについて報告する。その上で、今後ロボットに社会的感情を実装する上で必要なことを議論する。
%目的と手法がごっちゃになっているから分けて記載する？「そこで、我々はロボットに社会的感情を実装するための足掛かりとして社会的感情とロボットについて調査しまとめる。（目的）→具体的には心理学や…（手法）」
%主な調査結果を載せた方がいいとコメントされているので、「調査結果の一つとして，社会的感情の実装には自己他者認知のプロセスが重要であることが示唆された」とかかなぁ… →直さない
This study reviews research on social emotions in robotics. 
In robotics, the study of emotions has been pursued for a long time, including the study of their recognition, expression, and computational modeling of the basic mechanisms which underlie them. 
Research has advanced according to well-known psychological findings, such as category and dimension theories. 
Many studies have been based on these basic theories, addressing only basic emotions. 
However, social emotions, also referred to as higher-level emotions, have been studied in psychology. 
We believe that these higher-level emotions are worth pursuing in robotics for next-generation, socially aware robots. 
In this review paper, we summarize the findings on social emotions in psychology and neuroscience, along with a survey of the studies on social emotions in robotics that have been conducted to date. 
Thereafter, research directions toward the implementation of social emotions in robots are discussed. 

\end{abstract}


\section{Introduction}
%%%%%モチベーション
%感情の研究は大切
%身体とのつながり->ロボット
%基本的な感情はだいぶわかってきた
%これから先のことを考えると社会的感情を考えることが重要だろう
%%%%この論文の流れ
%本論文はロボットやagentにおける感情研究において、社会的感情に着目し、まとめたものである。
%社会的感情を扱う研究は現時点では非常に少なく、多くの研究が基本的感情に留まっている。
%また、感情に関する定義も各分野ではそれなりに固まってきていると感じるが、社会的感情に関しては、まだ統一的な定義が存在しない。
%本論文では社会的感情に関する定義や心理学的、神経科学的知見をまとめつつ、ロボットにおいて社会的感情の研究がどのように行われているかについて言及していく。
%そして、これからのロボティクスにおける社会的感情研究の課題について議論する。
%%%%%%サーベイに関する関連論文とこのサーベイ論文の主眼
%既存の感情とロボットに関するサーベイ論文としてSocial robotと感情についてのサーベイ論文\cite{cavallo2018emotion}や強化学習に焦点を置いたMotherlindらの論文などが存在する\cite{Moerland2018-kk}。
%これらは社会的感情について述べているものではないが、従来の感情とロボットの研究について網羅的にまとめているため、比較として本文中でも参照する。
%ただし、これらの論文は感情自体を学習するという立場ではないことに注意が必要である。
%心理学においては、感情が分化するという考えは古くからあり、ルイスは感情の発達をモデル化している\cite{lewis1995embarrassment}。
%最近の感情研究では、固定された基本的感情が存在するわけではなく、その基本は身体を基盤とした推論であり、発達や学習が重要な役割を果たしていると考えられている[リサ本、守口]。
%現状、感情研究については色々なことが明らかになっており、例えばinsulaが重要な役割を果たしていることや内受容感覚が感情の核になっていることが示唆されている。
%これらの研究の発展により、感情モデルの研究も様々進展している。
%しかしながらロボティクスにおける感情は、いまだに作り込みであったり、感情をはじめからあるものとして扱っている論文が大部分であり、発達や学習に着目している研究は少ない。

%今後のロボット研究において重要なのは、より複雑な社会的感情を扱うことであると考えるが、そのためには、感情が固定的で初めからあるものであると考えるのではなく、学習/発達するものであると捉える必要があると考える。
%これは、複雑な感情を網羅的にデザインすることが困難であるためである。
%また、社会的感情を含めた感情のメカニズムをロボットを通して解明する構成的アプローチを推進するためにもこうした点が重要である。
%議論のパートでは、社会的感情と基本感情を比較することで、発達について考察する。
%%%%

Emotion, also known as emotional intelligence \cite{emotionalIntelligenvce1996}, is an important element that forms the basis of human intelligence. 
Therefore, elucidation of its mechanisms and application to robots is an important direction of research. 
In particular, because of the connection between emotions and the body, emotion research using robots has attracted increasing attention \cite{damasio2019}. 
However, currently, emotional research in robotics has focused on basic emotions. 
It is necessary to broaden that focus to include social emotions to  create robots that are able to socially coexist with humans in the future, as well as to elucidate more complex emotional mechanisms. 

This survey paper focuses on social emotions in robot/agent research. 
There have been few studies focusing on social emotions, but many studies on basic emotions in robotics. 
Moreover, although the understanding of basic emotions has deepened considerably, our conception of social emotions is inadequate, and no unified definition of them exists, even in emotion studies. 
In this paper, we first overview the problems regarding the current study of emotions in robotics, followed by a presentation of the general theories of emotions. 
We then summarize the definitions of social emotions and review the associated psychological and neurological findings. 
Subsequently, we catalog how social emotions have been studied in robots. 
Finally, the challenges of social emotion research in robotics in the future are discussed. 

Previous research has attempted to survey and review the study of emotions and robots. 
\cite{cavallo2018emotion} reviewed emotions in social robots, whereas Motherlind {\it et al.} focused on reinforcement learning in emotion models \cite{Moerland2018-kk}. 
Although these projects did not describe social emotions, they comprehensively summarized emotion research in robotics; thus, we will refer to them in the text for comparison. 
Note that these survey papers did not provide a perspective on the learning and development of emotions themselves. 

In psychology, the idea of emotional differentiation has existed for decades, for instance, the model for emotional differentiation proposed by Lewis \cite{lewis1995embarrassment}. 
Recent emotional studies have shown that there are no fixed basic emotions. 
Instead, it is believed that the basis of emotion is inference based on our own body, and development and learning play an important role in the inference process \cite{LisaBook,moriguchi2013neuroimaging}. 
%At present, various things have been clarified about emotional research, and it is suggested that, for example, insula plays an important role and the sense of internal receptivity is the core of emotions.
As this area of research has progressed, various emotion model studies have been developed. 
However, most studies in robotics treat emotions as if they exist in their full form from the robot's beginning, and few studies have focused on the development of emotions in robots. 

%The author's interest lies in its developmental part, and I would like to consider development by summarizing social emotions and comparing them with basic emotions. 
We believe that it is important to address more complex social emotions in future robotic research. 
Therefore, it is necessary to perceive emotions as things which can develop and change, rather than being fixed and pre-existing.
This is because it is difficult to comprehensively design complex emotions. 
These points are also important for promoting a constructive approach to elucidate underlying emotional mechanisms, including social emotions, through robots. 
In the discussion section, we consider the development of social emotions by comparing them with basic emotions.

\section{Direction of this paper and current status of emotion studies in robotics}
\subsection{Overview of this survey paper}
%表に関する説明
%感情に関する研究は多様であり，その全てを網羅することは現実的ではない．
%ここでは，本論文が主眼とするロボットにおける社会的感情研究の目的や方法論を整理する．
%そして現状の問題点や，そうした問題を克服するために議論する視点について述べる．
Research on emotions is diverse, and it is not realistic to attempt to address them all in a single research paper. 
Here, we summarize the objectives and methodologies of social emotion research in robots, which is the main focus of this paper. 
Then, we describe the current problems and the viewpoints to overcome them.

\begin{figure}[t]
  \begin{center}
    \includegraphics[width=12.0cm]{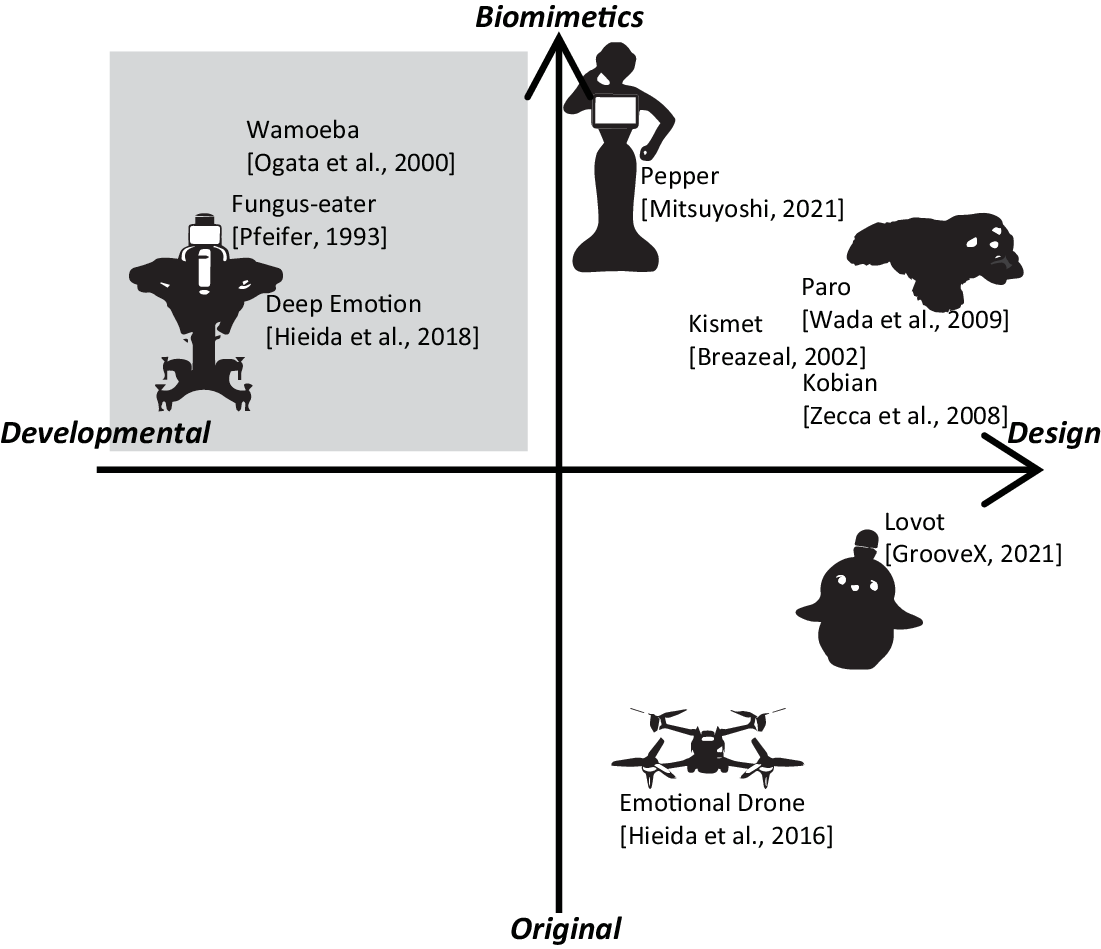}
    \caption{Approach to robot's emotions \cite{ogata2000development,pfeifer1993emotions,hieida2018deep,Mitsuyoshi2021pepper,wada2009long,breazeal2002designing,zecca2008design,lovot,hieida2016action}}
    \label{fig:intro}
  \end{center}
\end{figure}

%研究の目的や問題，視点の整理を表\ref{fig:intro}に示す．
%まず，研究の目的としては人に寄り添うためのパートナーロボットを創るということが挙げられる．
%これは工学応用的な目的である．
%そして，感情の本質を理解することに焦点を当てた科学的な目的もある．
%これはいわゆる構成論的アプローチと呼ばれるものである．
Table \ref{fig:intro} summarizes the purposes, problems, and viewpoints of the research on emotions in robotics. 
The first purpose of the research is to create a partner robot which can be emotionally close to people through use of the engineering application. 
Additionally, there is the scientific purpose, which is focused on understanding the essence of emotions.
This is the so-called ``constructive approach.''

%こうした目的に対するアプローチとして，大きく三つが考えられる．
%一つは，ロボットに人手でデザインするアプローチである．
%二つ目は，Developmentalな実現であり，人間が社会的な感情を生得的に持っているわけではなく獲得するという基本的な考え方に基づいて，ボトムアップに学習する枠組みを実現するものである．
%三つ目は，知見を得るための人を対象とした実験室実験である．
%三つ目に関しては，ロボット研究における社会的感情という文脈からは外れるが，社会的感情とは何かを知るための手段として重要であり，本論文でも社会的感情とは何かを考える上でまずこうした研究を参照する，そして，構成論的な研究という意味では，得られた知見の応用や，ロボットを使った実験も考えることができ，特にDevelopmentalなアプローチとは深い関係にある．
%つまり，これらのアプローチは完全に独立なわけではない．
There are three major approaches related to this goal.
The first is an approach which aims to manually design emotions using robots.
The second is the developmental approach, which emphasizes a bottom-up learning framework that is based on the basic idea that human beings do not have social emotions innately, but acquire them over the course of their lives. 
The third approach is using laboratory experiments with humans to obtain knowledge about social emotions. 
Regarding the third approach, although it is out of the context of social emotions in robotics research, it is important as a means to understand the key aspects of social emotions. 
We refer to these laboratory studies when considering the actual definitions of social emotions.

%\begin{table}[t]
%    \caption{Overview of this survey paper. Purpose, approach, amount of existing research, handling in this paper, issues, and perspectives are summarized in this table.}
%    \label{fig:intro}
%    \includegraphics[width=14.0cm]{./Fig/overview_table.eps}
    %社会的構成や言語の要素が表にないので入れる
%\end{table}
\begin{landscape}
\begin{table}[t]
 \caption{Overview of this survey paper. The purpose, approach, amount of existing research, how the papers were handled in this paper, issues, and perspectives are summarized in this table. Each number represents the corresponding description.}
  \label{table:overview}
  \centering
  \scriptsize
  \begin{tabular}{|c|c|c|c|c|c|} \hline
    \bf{Purpose} & \bf{Approach} & 
    \begin{minipage}{15mm}
    \bf{Existing Research} 
    \end{minipage}
    &
    \begin{minipage}{35mm}
    \bf{How the papers were handled in this paper}
    \end{minipage}
    & \bf{Issues} & \bf{Perspective} \\ \hline
    \hline
    %& & & & & \\
    \begin{tabular}{c}
    Building \\ partner \\ robot \\ (engineering)
    \end{tabular}
    & Design & Not many & 
    \begin{tabular}{c}
    Something over \\ basic emotions \\ ({\bf 5})
    \end{tabular} &
    %design issues
     \begin{minipage}{40mm}
     \begin{itemize}
        \setlength{\leftskip}{-7mm}
        \item %そもそも社会的感情が何か分からないのでデザインすることが困難
        It is difficult to design, because the definition of social emotions is not clear.
        \item %作る必要があるのかが明らかではない
        Need for implementation of social emotions in robots is not clear.
        %\item 認識と表出が分離している
        \end{itemize}
     \end{minipage}
    & 
    %design perspective
    \begin{minipage}{55mm}
     \begin{itemize}
        \setlength{\leftskip}{-7mm}
        \item %社会的感情とは何かを本サーベイで整理した
        This survey organized what social emotions are. ({\bf 4}) 
        \item %社会的文脈の中で振る舞うロボットには必要である
        Robots that behave in a social context need to have social emotions. ({\bf 6.1, 6.2}) 
        \item %インタラクションのレベルによってはデザインすることができる
        Social emotions can be designed depending on the level of interaction. ({\bf 6.2})
        \item %人間の持ち得る全てをデザインすることは困難
        It is not plausible to design all social emotions that humans have. ({\bf 6.2}) 
        \item %デザインに際しては社会構成主義の知見が参考になる
        When designing social emotions, the knowledge of social constructivists is helpful. ({\bf 6.2})
        \end{itemize}
     \end{minipage}
    \\  \cline{2-6}
    & Developmental & Few & 
    \begin{tabular}{c} Core affect \\ ({\bf 6.5}) 
    \end{tabular} 
    &
    %developmental issues
    %\footnotesize{
    %\begin{tabular}{c}
    %色々と複雑で難しい
    %\end{tabular}
    %}
    \begin{minipage}{40mm}
        \begin{itemize}
            \setlength{\leftskip}{-7mm}
            \item %発達メカニズムは複雑で難しい
            Emotional developmental mechanisms are complex and difficult to model. 
            \item %基本メカニズムはわかりつつあるが何が足りないいかが不明
            The basic mechanism is understood, but the limitations are unclear.
        \end{itemize}
    \end{minipage}
    &
    %developmental perspective
    %\footnotesize{
    %\begin{tabular}{c}
    %発達ロボティクスや \\ 記号創発ロボティクスの考え方が重要
    %\end{tabular}
    %}
    \begin{minipage}{55mm}
        \begin{itemize}
            \setlength{\leftskip}{-7mm}
            \item %自身の身体を通して真に社会的感情を理解できるロボットを創るためには分化の考え方が重要
            The idea of differentiation is an important prerequisite to create a robot that can truly understand social emotions through its own body. ({\bf 6.3})
            \item %現在の機械学習の発展を利用できる可能性がある
            May take advantage of current machine learning developments. ({\bf 6.4})
            \item %発達ロボティクスや記号創発ロボティクスの考え方が重要
            The idea of developmental robotics and symbol emergence in robotics is important. ({\bf 6.6})
        \end{itemize}
    \end{minipage}
    \\ 
    \cline{1-1}
    \begin{tabular}{c}
    Understanding \\ underlying \\ mechanism \\  (science)\\
    \end{tabular}
    &  &  &  &  &
     \begin{minipage}{55mm}
        \begin{itemize}
            \setlength{\leftskip}{-7mm}
            \item %身体としてはソフトロボティクスに期待することができるかもしれない
            As a research body, we may be able to expect soft robotics. ({\bf 6.5}) 
            \item %社会構成主義的な要素を陽に取り入れることも今後の重要な課題
            Appropriately incorporating a social constructivist view is an important issue for the future. ({\bf 6.5}, {\bf 6.6})
            \item
            %ロボットのモラル行動をモラル感情によって作り出せるかもしれない
             It may be possible to make a robot moral and for robots to understand morals by implementing social  emotions. ({\bf 6.7})
        \end{itemize}
    \end{minipage}
    \\ 
    \cline{2-6}
    %\begin{tabular}{c}
    %Understanding \\ underlying \\ mechanism \\  %(science)\\
    %\end{tabular}
    & Experimental & Many & \begin{tabular}{c} Organizing findings \\ ({\bf3}, {\bf4}) \end{tabular} &
    %experimental issues
    %\footnotesize{
    %\begin{tabular}{c}
    %様々な知見の融合が難しい
    %\end{tabular}
    %}
    \begin{minipage}{40mm}
        \begin{itemize}
            \setlength{\leftskip}{-7mm}
            \item %様々な知見の融合が難しい
            It is difficult to integrate various domains of knowledge. 
            \item %特に社会的な側面と個体の側面が乖離している
            In particular, the social and the individual aspects are distinct. 
        \end{itemize}
    \end{minipage}
    &
    \begin{minipage}{55mm}
        \begin{itemize}
            \setlength{\leftskip}{-7mm}
            \item %様々な知見を融合させるためにもロボットによる構成的な研究が重要になる
            Constructive research in social emotions using robots is important for fusing various findings. ({\bf 6.6})
            \item %構成的な研究の知見を科学的な研究にフィードバックする必要がある
            It is necessary to feed the findings of the constructive approach back into scientific research. ({\bf 6.6})
        \end{itemize}
    \end{minipage}
    \\ \hline
  \end{tabular}
\end{table}
\end{landscape}
%%%table contents%%%

With respect to the constructive approach, it is possible to consider the utilization of the obtained knowledge and experiments using robots to deepen that knowledge. 
In particular, it is closely related to the developmental approach. 
In other words, these approaches are not completely independent. 

%後の章で詳しく見ていくが，こうした内容の既存研究はどれほど存在するであろうか．
%実際，ロボットに社会的感情をデザインした研究はあまり存在しない．
%さらに，ロボットが社会的感情を獲得するような発達を基盤にした実例はほとんど見当たらない．
%一方で，人を対象として社会的感情を調べる実験的な研究は数多く存在する．
%これらの具体的な内容については，後の章で詳しく述べる．
In a later chapter, we will discuss the numerous studies pertain to these different aspects of social emotions. 
In fact, there are few studies that have designed social emotions for robots. 
Furthermore, there are few examples based on developmental research, wherein robots are able to acquire social emotions. 
Alternatively, there are many experimental studies that have examined social emotions in humans. 
The specific contents of these will be described in detail in a later chapter. 

%こうした状況を踏まえて，著者らが考える問題点を表に示しているが，これらの問題について，本論文ではサーベイ結果に基づき議論する．
Based on this evaluation of the state of emotion research in robots, the problems that the authors consider are shown in the table, and these problems will be discussed based on the survey results in this paper. 

\subsection{Emotion studies using existing robots}
\label{sec:pre_robot}
%図について追記
%現存する実際のソーシャルロボットにおける感情の実装について、主要なものを図\ref{fig:intro}にまとめた。
%ただしこの図における各ロボットの配置は、著者の主観によるものである。
%この図では横軸が感情の実装がデザインであるか、感情理解を目指した構成論的なモデルであるかを表し、縦軸が既存の生物に似せて作られているか、オリジナルな表現であるかを表す。感情理解を目指すものに関しては、完全にオリジナルではその目的を果たすのが困難であるため、原理的に第3象限には該当する研究はない。デザインによる研究は製品なども多く、安定し、望んだ振る舞いが必要とされているものになっている。
%我々はこの図の中でグレーで表される第2象限のエリアこそ、これからのロボットの感情研究において重要と考えており、人のような複雑な感情をロボットが表現するのにデザインだけでは難しいと考えている。
%当然、ロボットがどのような場面で使用されるかに依存する部分もあるが、ロボットが人と友人のような関係を築く未来を想像するのならば第2象限で表される研究は必須である。

Some implementations of emotions in existing social robots are summarized in Fig. \ref{fig:intro}. 
However, all of the robots in this figure are famous, and their arrangement is based on the author's subjective assessments. 
In this figure, the horizontal axis indicates whether the implementation of emotions was a design or a developmental model. 
%aimed at understanding emotions.
The vertical axis indicates whether the robot was made to resemble an existing creature or was it an original expression. 
It has been difficult to achieve a developmental model with a completely original robot; hence, there is no research that corresponds to the third quadrant. 
We believe that the area in the second quadrant, which is shown in gray in this figure, is important for future research on robot emotions, as it is difficult for robots to express complex emotions like humans by design alone. 
Although it depends in part on the context in which the robot is used, the research represented by the second quadrant is indispensable if one imagines a future in which robots will establish friend-like relationships with humans.

\section{General theory of emotions}
\subsection{Definition of emotions}
Considering the definition of emotion, to the best of our knowledge, there is no universal consensus on the definition of the term ``emotion.'' 
Damasio, for example, distinguished between emotions and feelings. 
Emotions were defined as a series of physical reactions, changes in the state of internal organs and skeletal muscles, and changes in internal states. 
Feelings, alternatively, were defined as the recognition of emotional states \cite{damasio2003looking}. 
Other definitions of emotions are based on unconsciousness and consciousness, transient and subjective experiences, emotions that are suddenly caused by stimuli, etc. 
The word ``affect'' is occasionally used to refer to all emotions, and it has a wide variety of meanings. 
In this paper, all items are represented using the word ``emotion'' without dividing the concept into a further detailed meaning. 

\subsection{Psychological theories on emotions}
%感情理論の円環モデルと基本感情など追記
There are several different psychological theories of emotion. There are category theories, which regard emotions as categories, and dimensional theories, which represents emotions in an emotional space composed of several dimensions. 
Ekman's six common emotions (anger, disgust, fear, happiness, sadness, surprise, and neutral) which exist regardless of culture is an example of category theory \cite{ekman}. 
Although subsequent research has disputed this claim, it is still a popular theory. 
Russell's circumplex model, conversely, is well-known as a dimensional theory of emotion. 
Russell stated that emotions are placed in a two-dimensional space consisting of valence and arousal (VA) axes \cite{Russell}. 

Other examples of extensions of the two-dimensional Russell's circumplex model are pleasure-arousal-dominance (PAD) \cite{mehrabian1974approach} and valence-arousal-stance (VAS) \cite{breazeal1999build}. 
In PAD, the D dimension indicates dominance-submissiveness, which represents the control state of the individual emotion relative to particular situations and influences from others. 
In VAS, the stance specifies how approachable the percept is to the agent.
The stance dimension indicates an axis ranging from an open stance to a closed stance. 

Furthermore, Plutchik's wheel of emotions can be regarded as both a dimensional theory and a category theory \cite{Plutchik}. 
As a dimensional theory, it is represented by a cone-like shape, wherein the emotional intensity becomes stronger toward the bottom surface of the cone and weaker toward the apex. 
Moreover, the opposite side of the cone expresses the opposite emotions. 
As a categorical theory, this cone-shaped model is composed mainly of eight basic emotions, and through their combination various other emotions can be expressed.

%Neuroscientific research
\subsection{Neuroscience studies}
Recent studies have clarified the importance of the body in emotions, as previously suggested by William James through the peripheral theory of emotions \cite{james}. 
In recent studies in cognitive neuroscience, it has been reported that the perception of the internal state, also known as interoception, is the key to the subjective experience of emotions \cite{Terasawa2013humanbrainmapping}. 
According to the emotional quartet theory, a brainstem-centered system corresponds to an emotional system \cite{KOELSCH20151}. 
The brainstem is the oldest brain structure, and the reticular formation plays an important role in this brainstem-centered system.
Another important aspect of the relationship between emotions and the body is Damasio's somatic marker hypothesis, which assumes that emotions efficiently evaluate external stimuli through our own body \cite{Damasio}. 

As previously mentioned, the body is the origin of emotions and is thus indispensable to the study of emotions. 
It is believed that basic emotions, such as anger, joy, disgust, fear, sadness, and surprise, exist regardless of culture, even though how these emotions are expressed might be different \cite{ekman}. 
This might be because humans share a similar body and environment, which supports the notion that emotions are based on our physical body and environment. 
The idea of active inference is also related to the body system, such as visual attention, which is related to the evaluation of external stimuli \cite{Friston2010,Seth20160007}. 

Note that emotions are related to decision-making and causal reasoning \cite{Ledoux}. 
For example, misattribution of a physical reaction, known as the suspension bridge effect, is the false attribution of one's personal-physical reaction to another person by experiencing the fear of one's self is in danger while meeting someone \cite{Dutton}. 
This higher-level cognitive process is believed to be closely related to the system centered on the orbitofrontal cortex, as well as to the reinforcement learning module derived from the corticobasal loop.
The relationship between active inference and reinforcement learning is discussed in \cite{Friston2009PLOS}. 

Memory-based systems are also considered to be important components of emotions. 
In the quartet theory, the hippocampal-centered system, which mainly involves the hippocampus and the amygdala, are the important brain structures involved in emotion \cite{KOELSCH20151}. 
Amygdala activity during the experience of emotions is important, and it has extensively been studied. 
For example, the well-known limbic circuit Yakovlev includes amygdala \cite{YAKOVLEV1948}. 
Similarly, the Papez circuit is also a well-known limbic circuit that includes the hippocampus \cite{papez1937}. 
Although these circuits are independent, they interact with each other through the cortex, basal ganglia, and diencephalon, and are closely related \cite{mendoza2007clinical}. 

As described above, emotions need to be considered as a network rather than as a localized circuit, because each function is interrelated and several models have been proposed for this integration. 
In Damasio's conceptual model, the amygdala evaluates the information, the hypothalamus induces a physical reaction, and then the physical reaction actually occurs. This is the emotional state. 
The information is sent to the cerebral cortex as an internal receptive sensation, integrated with external receptive sensations, and emotional feelings are perceived \cite{Damasio}. 
In the model of Moriguchi and Komaki, a core affect (i.e., emotional state) is formed based on the interoception representing the physical state of the physical body \cite{moriguchi2013neuroimaging}. 
Emotions are perceived by integrating and categorizing core affects with information such as contexts and concepts. 
This is a model that was proposed based on the alexithymia neuroimaging research.

\subsection{Social constructivist view of emotions}
\label{sec:social}
%人がどのような感情を持つかは，社会的な要因が大きく関係している．
%感情のsocial constructivistsは，社会に感情の根源を見出そうとする．
%社会的構築主義の走りはAverillであり，感情を完全に理解するためには，社会レベルでの解析が必要であることを主張した \cite{AVERILL1980305}．
%より最近では，Boigerらが感情を科学するために，social worldを考慮することの重要性を強調している \cite{Boiger2012}．
%感情は，非常に複雑なプロセスであり，生物学的な制約に基づいているが，その個体が発達する社会の影響を考慮しなければ完全に理解することはできない．
%彼はさらに，感情はオンゴーイングでダイナミックで、インタラクティブなプロセスであり，3つの文脈，つまりは，in-the-moment interactions, relationships, and culturalが埋め込まれていると述べている \cite{Boiger2012int}．
Social factors have a great deal to do with how people feel.
The social constructivists in emotions seek to find the source of emotions in society.
The originator of social constructivists in emotions was Averill, who argued that a social level analysis was needed to fully understand emotions \cite{AVERILL1980305}. 
More recently, Boiger emphasized the importance of considering the social world for the science of emotions \cite{Boiger2012}. 
Emotion is a very complex process, based on biological constraints, but it cannot be fully understood without considering the social impact of the individual's development. 
He further stated that emotions are an ongoing, dynamic and interactive process that embeds three contexts: in-the-moment interactions, relationships, and culture \cite{Boiger2012int}.

%cultureとの関係性
%特に文化は，社会構成主義では重要な役割を果たす．
%感情と文化のかかわりに関しては，多くの研究がなされており，文化による感情の類似性や相違性など様々な議論がある \cite{cultualdifference1992,LIM2016105}．
%また，言語と感情の関わりも議論されている．
%この言語と感情の関わりの究極は，Harréの``Emotionology Principle''  \cite{emotionology2009} である．
%この考え方では，感情はそれを表現する言語がどのような条件において使われるかを特定することで理解できるというものである．
Culture in particular plays an important role for social constructivists. 
Much research has been conducted on the relationship between emotions and culture, and there have been various discussions about the similarities and differences of emotions between cultures \cite{cultualdifference1992,LIM2016105}.
The relationship between language and emotion has also been studied.
The ultimate relationship between language and emotion is Harré's ``Emotionology Principle'' \cite{emotionology2009}.
The idea is that emotions can be understood by identifying under which conditions the language that expresses these emotions is used.

The perspective that emotions are socially constructed is important, but it would be an extreme view to think that it represents the totality of emotion formation. 
Emotional mechanisms exist in the cognitive domain, based on the human body as described earlier. 
This is also a premise in claims of recent social constructivists. 
\cite{ITRW2000} summarized the social constructivist perspective and discussed how emotions could be studied under the fusion of various views, such as cognitive and psychological perspectives. 
This integrated perspective seems to be important to fully understand emotions. 
\cite{Aranguren2016} stated that there is a need to separate ontology and methodology in emotional research. 
This is based on the assumption that ontological emotions that are described by language can lead to oversights regarding the nature of emotions. 
This is also suggested by emotional studies in nonhuman primates, who lack linguistic abilities. 

%ここは議論に移動したほうがよいかもしれない
%社会構成主義的な主張は，ロボットを用いた感情研究に対しても当てはまる．
%そもそも，ロボットに感情をデザインすることは，感情の社会的な機能を実現しようとしてると捉えることができる．
%これはまさに社会構成主義的な発想であり，そうしたデザインは社会構成主義を突き詰めていく必要がある．
%現状のロボットにおける感情は，ある種の``Emotionology Principle''に基づいており，より汎用的な感情表現をデザインするためには、オントロジーとメソドロジーの分離が必要であると思われる．
%Social constructivist claims also apply to emotional research using robots.
%In the first place, designing emotions on a robot can be regarded as trying to realize the social function of emotions.
%This is exactly the iedea of social constructivist, and such a design needs to pursue social constructivist view of emotions. 
%Emotions in current robots are based on a kind of ``Emotionology Principle,'' and it seems necessary to separate ontology and methodology to design more general-purpose emotional expressions. 

%angry ですら社会的なsophisticated な感情
%社会構成主義的な感情理論のもう一つの特性は，基本的な感情と社会的感情を陽には分けないところにある．
%これは，感情が社会的に構築されると考えれば自然なことである．
%例えば，angryはそれ自体sophisticatedな社会的感情であるとみなす．
%本論文では，次章で社会的感情について述べるが，そこでは社会的構築の視点には触れない．
%しかし本節で述べた社会的構築の要素が、社会感情の構築に含まれることは明らかである．
An important characteristic of the social constructivist's emotion theory is that it does not explicitly separate basic emotions from social emotions. 
This is a natural consequence of the idea that emotions are socially constructed. 
For example, anger is considered to be a sophisticated social emotion in itself. 
In this paper, we will discuss social emotions in the following chapter, but we will not touch on the perspective of social constructivist. 
However, it is clear that the elements of social construction described in this section should be included in the construction of social emotions. 

%\section{社会的感情の定義}
%ここまでは感情全般に関して述べられている定義や理論に言及した。ここからは社会的感情の定義について述べる。Parkinsonらはsocial emotions として、下記の感情を挙げている\cite{parkinson2005emotion}。
\section{What are social emotions?}
\label{sec_def}
\subsection{Taxonomic definition of social emotions}
Until now, we have established general definitions and theories about emotions. 
From here, we focus on the definition of social emotions. 
Parkinson {\it et al.} listed the following as social emotions \cite{parkinson2005emotion}. 
\begin{itemize}
    \item {\bf Embarrassment}; this is caused by unwanted interpersonal attention, and serves to deflect it.
    \item {\bf Shame}; this is simply a stronger and more enduring form of embarrassment.
    \item {\bf Guilt}; this tends to focus on a particular incident, rather than a more general moral failing. Further, it is a response to accusations from a close social partner.
    \item {\bf Jealousy and Envy}; both refer to negative reactions to the good fortune of another person.
    \item {\bf Love}; this is a statement about the perceived status of a relationship or a promise of commitment. Moreover, it is quite diverse in character: sometimes positive and exciting, sometimes painful, and sometimes calm and serene.
    \item {\bf Fago}; this represents concern for other people when they are alone, ill, or, by contrast, exhibiting qualities of interpersonal sensitivity and social intelligence.
    \item {\bf Grief}; this is a feeling of suffering. For example, the experience resulting from the death of someone close.
\end{itemize}

%一方で、2008年のhareliとparkinsonの論文では、上記とは異なる定義として、社会的な関係を有するものは全て社会的感情と扱っており、同様の感情であってもsocialとnon-socialがあると述べている\cite{hareli2008s}。
%また、Jon Elsterは自分自身の評価/他人の評価、行動の評価/性格の評価、正の評価/負の評価の３組の特徴の組み合わせから８つの社会的感情を提案している\cite{elster1999strong}(table \ref{table:elster})。
%本定義では、従来基本感情とされていたものも社会的感情に含まれることになる。Parkinsonらの従来の社会的感情のリストとは、shameとguiltなどが共通している。
%
Hareli and Parkinson defined anything with a social relationship as a social emotion, which is different from the above definition. 
They stated that both social and non-social perspectives exist, even with similar feelings \cite{hareli2008s}. 
In addition, Elster proposed eight social emotions from a combination of three types of characteristics: evaluation of oneself or evaluation of others, evaluation of behavior/evaluation of personality, and evaluation of positive/negative, as shown in Table \ref{table:elster} \cite{elster1999strong}. 
According to this definition, social emotions include those that were traditionally considered as basic emotions. 
Shame and guilt are common between Parkinson's social emotion category and Elster's social emotion category. 

%\begin{itemize}
%    \item Shame
%    \item Contempt
%    \item Guilt
%    \item Anger
%    \item Pridefulness
%    \item Liking
%    \item Pride
%    \item Admiration
%\end{itemize}
\begin{table}[t]
 \caption{Definition of social emotions of Elster \cite{elster1999strong}}
  \label{table:elster}
  \centering
  \begin{tabular}{|l|c|c|c|c|} \hline
    \bf{Evaluation} & \multicolumn{2}{|c|}{\bf{Behavior}} & \multicolumn{2}{|c|}{\bf{Personality}}   \\ \cline{2-5} 
    & \bf{Positive} & \bf{Negative} & \bf{Positive} & \bf{Negative} \\ \hline
    \bf{Self} & Pride & Guilt & Pridefulness & Shame \\ \hline
    \bf{Others} & Admiration & Anger & Liking & Contempt / Hatred \\ \hline
  \end{tabular}
\end{table}

%また、social emotionのsub categoryとしてmoral emotion や self-conscious emotionが存在する\cite{hareli2008s}。Moral emotions are defined as emotions that are intrinsically linked to the interests or welfare of society as a whole or to persons other than the agent. 道徳的感情は対人関係の出来事の文脈での道徳的違反の認識によって容易に引き起こされ、道徳的行動を導く。このカテゴリーには、shame, guilt, regret, embarrassment, contempt, anger, disgust, gratitude, envy, jealousy, schadenfreude, admiration, sympathy and empathyが含まれる。
%一方、自己意識的な感情は、特定の出来事や状況が自己評価や福祉に影響を与えることに個人が気付いたときに生じる感情と見なされる。よく言及される例は、shame, guilt, pride, and embarrassmentである。
%また、Lewisは感情発達の理論モデルを提案しており、その中で自己意識的感情は２歳になるまで出現しないとしている\cite{lewis1995embarrassment}。これは、乳児が自分が他者と異なる存在であることを意識することがこれらの感情に関係しており、自己意識的感情と自己鏡映像の認知との関係を指摘している。感情を大別して非自己意識的感情と自己意識的感情の二つと定義する例も存在し、その場合は非自己意識的感情は基本感情と同一である。
%また、呼称としてLewisは一次的感情と二次的感情という大別を行っており、この際も一次的感情はいわゆる基本感情を指す。そのほかにも低次、高次感情という表現もある。

%いずれにしても、社会的感情に類する感情の分類の定義において、共通しているのは他者の存在である。自分と異なるが、自分と類する存在である他者とそれらが複数個体となって構成される社会が関係し、社会的感情が定義されている。

%次項からは、各分野において、どのような社会的感情についての研究があるかに触れていきたい。

Furthermore, moral emotions and self-conscious emotions are considered to be sub-categories of social emotions \cite{hareli2008s}. 
Moral emotions are defined as emotions that are intrinsically linked to the interests or welfare of society as a whole, or to persons other than the agent. Moral emotions are easily triggered by the perception of moral violations in the context of interpersonal events, which can lead to moral behavior. This category includes shame, guilt, regret, embarrassment, contempt, anger, disgust, gratitude, envy, jealousy, schadenfreude, admiration, sympathy, and empathy.
Self-conscious emotions are emotions that occur when an individual becomes aware that a particular event or situation affects their own self-esteem or well-being. Common examples include shame, guilt, pride, and embarrassment. 
Lewis, who proposed a theoretical model of emotional development, claimed that self-conscious emotions do not appear until the age of two \cite{lewis1995embarrassment}. 
This indicates that infants' awareness of being different from others is related to these emotions. Furthermore, Lewis suggested a relationship between self-conscious emotions and the recognition of themselves in a mirror. 

Emotions can also be classified into non-self-conscious emotions and self-conscious emotions wherein the non-self-conscious emotions are the same as the basic emotions. 
In addition, Lewis divided emotions into primary and secondary wherein the primary emotions refer to basic emotions. 
There are also expressions, such as lower-level and higher-level emotions.

%In any case, the existence of others is common in the definition of emotion classification, similar to social emotions. 
Regardless, the existence of other people is common in the aforementioned definitions of social emotions. 
Social emotions are thus defined by the relationship between others, who are different from oneself but in other ways similar to oneself, and the society, which comprises multiple individuals.

\subsection{Social emotions and psychology}
In psychological research on social emotions, guilt and shame/embarrassment are mainly studied because they have important functionalities for maintaining our social groups. 

Niedenthal {\it et al.} summarized the emotions that can occur within a group \cite{niedenthal2012social}. 
The authors divided the emotions within the group into group emotions and group-based emotions. 
The group emotions consisted of emotions that occur as a function of being in a group, such as sharing joy, whereas group-based emotions consisted of emotions that occur from a group, such as individual guilt. 
In particular, group-based emotions can be used as a predictor of group behavior against social injustices.

According to Hoffman, guilt can be defined as a strong contempt for oneself to injure others unfairly. 
For instance, based on the parental involvement, their child pays attention to the victim's pain and can recognize and empathize with their situation. 
Hoffman stated that guilt can cause empathic distress and foster empathy-based guilt \cite{hoffman2001empathy}. 
In addition, the guilty conscience that arises by collectively belonging to an ethnic group, nation, group, etc., rather than the act of the individual itself, is called collective guilt \cite{manstead2004collective}. 

Shame/embarrassment is an unpleasant emotion, and in some cases, it is one of the emotions that can completely disrupt interpersonal interactions and strongly afflict an individual for long time \cite{miller2007embarrassment}. 
Shame itself is an emotion with a variety of meanings and is attributed to social evaluation concerns, self-image disagreements, interaction confusion, and reduced self-esteem \cite{higuchi2008comparison}.

Comparing oneself with others is called making a ``social comparison.'' 
If jealousy and envy are caused by the awareness of things which an individual sees but does not have, it is considered that a social comparison is involved in the arousal process of these emotions. 
One of the social comparison theories is the self-evaluation maintenance (SEM) model \cite{tesser1984friendship}. 
This model assumes two processes, the comparison process and the reflection process. 
The comparison process is a comparison of the content you are interested in, and the reflection process is a comparison of the content that you are not interested in. 
Currently, jealousy and envy are considered to be related to the comparison process. The circumstance in which others are viewed as superior is the turning point in determining whether one experiences jealousy or envy. 

\begin{figure}[t]
  \begin{center}
    \includegraphics[width=14.0cm]{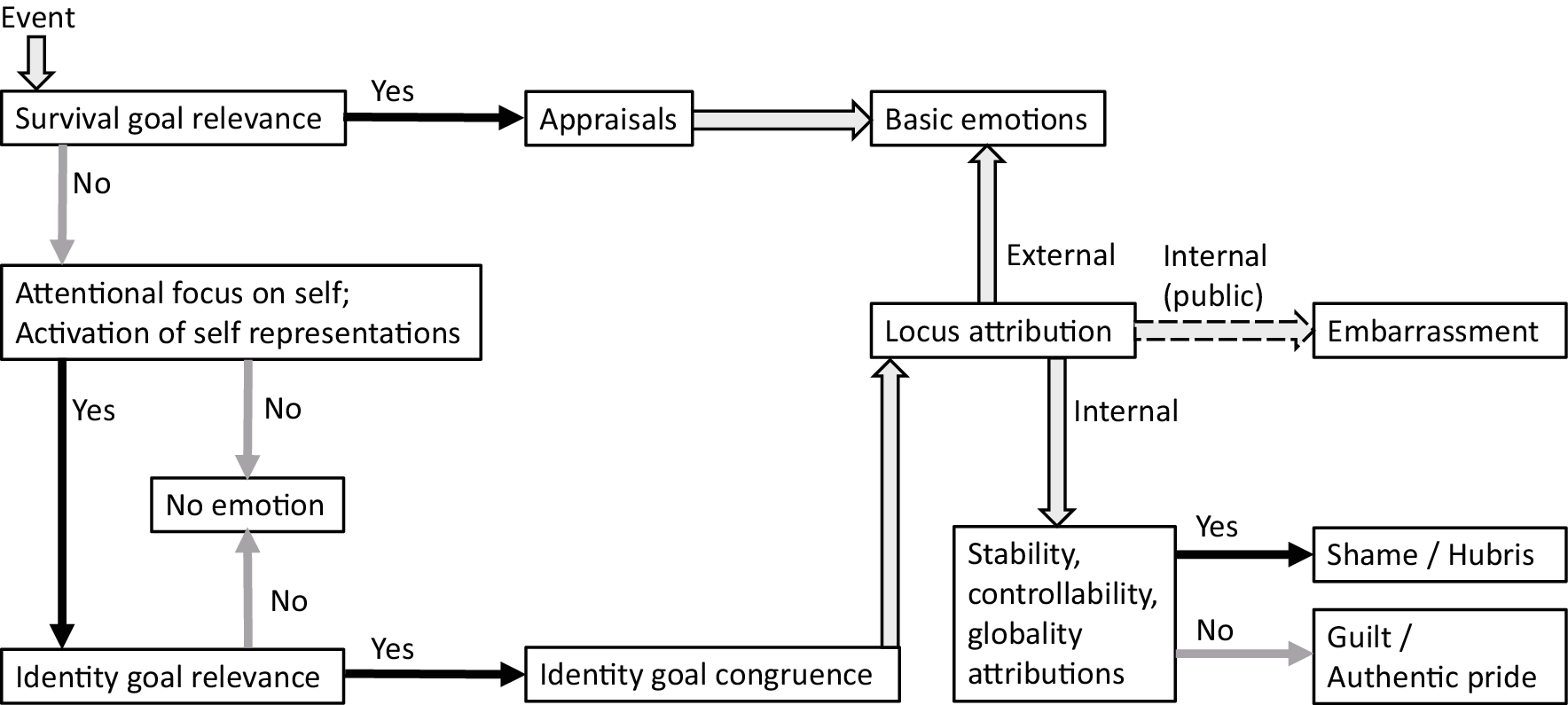}
    \caption{Process model of self-conscious emotions \cite{tracy2007self}}
    \label{fig:pride}
  \end{center}
\end{figure}

Pride is a positive self-conscious emotion that is experienced when one's behavior, remarks, and traits are superior or desirable, and are positively evaluated by others \cite{fischer1995self}. 
The process of experiencing pride, etc., is modeled by Tracy and Robins \cite{tracy2007self}, as illustrated in Figure \ref{fig:pride}.
Specifically, hubristic pride, such as shame, results from internal, stable, uncontrollable, and global attributions. Highly authentic pride, such as guilt, results from internal, unstable, controllable, and specific attributions. 

\begin{figure}[t]
  \begin{center}
    \includegraphics[width=13.0cm]{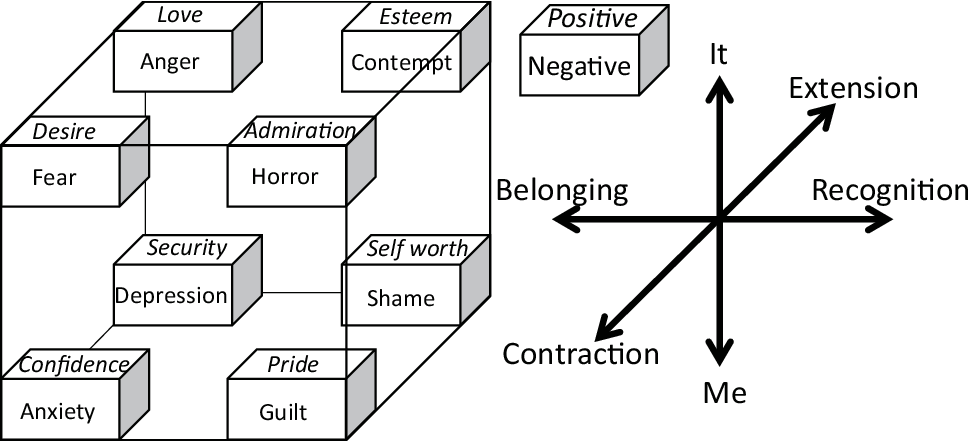}
    \caption{Structural theory of social emotions using four interpersonal factors \cite{de1986emotions}}
    \label{fig:itme}
  \end{center}
\end{figure}

%また、1986年にRivera and Grinkis がinterpersonal factorsによってstructural theory を提案している\cite{de1986emotions}(Figure \ref{fig:itme})。
%この提案のinterpersonal factorsは"It--Me", "Extension--Contraction", "Belonging--Recognition" and "Positive--Negative"で構成されている。
In 1986, Rivera and Grinkis proposed a structural theory based on interpersonal factors \cite{de1986emotions}, as shown in Figure \ref{fig:itme}. 
The interpersonal factors of this proposal consist of ``It--Me,'' ``Extension--Contraction,''  ``Belonging--Recognition'' and ``Positive--Negative.'' 
The ``It--Me'' factors refers to a choice of whether to direct a given emotion toward another person or toward yourself. 
The ``Extension'' factor reflects the distinction between accepting other people or rejecting them. 
The ``Contraction'' factor reflects the distinction between wanting to be associated with other people or withdrawing from them. 
The ``Belonging'' factor reflects whether the focal person belongs to another person or form a unit together, and the ``Recognition'' factor involves social recognition and comparison. 

\subsection{Social emotions and brain activity}
Koush {\it et al.} examined the brain activities associated with positive emotions in a social context using functional magnetic resonance imaging (fMRI) \cite{koush2019brain}. 
Their results demonstrated that self-referential positive-social emotion regulation recruited a distributed network of prefrontal, temporoparietal, and limbic brain areas. 

Immordino-Yang {\it et al.} investigated brain activities related to compassion \cite{immordino2011me}.
Using a compassion- and admiration-inducing procedure, they found that social emotions strongly recruited a neural region comprising the posterior/inferior precuneus and the neighboring retrosplenial cingulate, which are both involved in high-level associations of interoceptive information from visceral sensation and regulation. 
Functionally, this area is involved in the default network, which has been suggested to correspond to the feeling of self, personal salience, and higher-level consciousness.

Klapwijk {\it et al.} conducted an experiment wherein they read scenarios that evoked embarrassment and guilt (social emotions) and disgust and fear (basic emotions) to subjects and measured their brain activity using fMRI \cite{klapwijk2013increased}. 
Psychophysiological interaction (PPI) analysis revealed that the right posterior superior temporal sulcus (PSTS) and the right temporoparietal junction (TPJ) showed greater functional connectivity with the dorsomedial prefrontal cortex (DMPFC) during social emotions than with basic emotion. 
These brain regions were suggested to be involved in mentalizing and acquiring the viewpoints of others.

Grecucci {\it et al.} summarized the brain regions related to emotional regulation as a mechanistic study of social emotional regulation \cite{grecucci2015mechanisms} (Table \ref{table:part}). 
First, individual emotion regulation (IER) has been reported to be related to the dorsolateral prefrontal cortex (DLPFC), ventrolateral prefrontal cortex (VLPFC), anterior cingulate cortex (ACC), amygdala, striatum, and orbitofrontal cortex (OFC), among others. 
The DLPFC is generally believed to control attention and working memory. 
The ACC is involved in monitoring and controlling ongoing processes. The VLPFC appears to be responsible for choosing an appropriate response to a goal and suppressing inappropriate responses. 
The area of interest for reassessment is the amygdala, which is believed to be an important structure that supports the refinement of external and internal emotional and negative stimuli. 
In addition, the striatum and insula have little relationship with IER. 

Socially cued emotions are emotions that are caused by interactions with others, such as rejected sadness.
Socially cued emotion regulation (SER) involves different brain regions than IER, depending on the social context, SER can involve the medial prefrontal cortex (MPFC), the medial orbitofrontal cortex (MOFC), the posterior cingulate cortex (PCC), and the amygdala. 
In particular, the MPFC is related to the mentalizing of oneself and others, and the PCC is related to the attribution of others' emotions.
In addition, when emotions are downregulated using mentalization, subjects are shown to have weak emotional responses, less rejection behavior, and less neural activity when they received an unjust offer. 
This emotional modulation is observed on the insula, which has been found to represent the emotional experience of the viscera. 

Furthermore, interpersonal emotion regulation (I-PER) is a strategy for regulating the emotions of others during interactions.
The same brain regions are activated as in IER/SER, with the additional activation of other regions, as listed in Table \ref{table:part}.

\begin{table}[t]
 \caption{Key brain regions related to emotion regulation \cite{grecucci2015mechanisms}}
  \label{table:part}
  \centering
  \begin{tabular}{|l|l|} \hline
    \multicolumn{2}{|l|}{\bf{Individual Emotion Regulation}}    \\ \hline 
    \bf{Regulating Regions} & \bf{Regulated Regions}  \\ \hline
    DLPFC & Amygdala  \\
    VLPFC & Striatum \\
    ACC & OFC  \\ \hline
    \multicolumn{2}{|l|}{\bf{Socially Cued Emotion Regulation}}   \\ \hline 
    \bf{Regulating Regions} & \bf{Regulated Regions}  \\ \hline
    VMPFC & Insula  \\
    VLPFC & Striatum \\
    TPJ & Cingulate \\
    Temporal Pole &  \\
    ACC & \\ \hline
    \multicolumn{2}{|l|}{\bf{Interpersonal Emotion Regulation}}   \\ \hline 
    \multicolumn{2}{|l|}{\bf{Regulating Regions} } \\ \hline
    \multicolumn{2}{|l|}{Left Temporal Pole / Inferior Temporal Gyrus} \\
    \multicolumn{2}{|l|}{Rostral Medial Prefrontal Cortex}\\
    \multicolumn{2}{|l|}{Posterior Insula} \\
    \multicolumn{2}{|l|}{Cingulate Gyrus} \\
    \multicolumn{2}{|l|}{Bilateral Caudate} \\
    \multicolumn{2}{|l|}{Cuneus / Inferior Parietal Lobule} \\ \hline
  \end{tabular}
\end{table}

\section{Survey on social emotions in robotics}
\subsection{Survey based on survey papers of emotional robotics}
%まず初めに、ロボットに関連する感情の調査論文について言及する。
%Yanらはsocial robotでのemotional spaceについて調査している\cite{Yan2021-zf}。
%cavalloはsocial robotでの感情について調査し、ekmanの基本感情を使用する例が多いことを示している\cite{cavallo2018emotion}.
%Moerlandは強化学習をターゲットとしてrobotやagentでのemotion model研究についてまとめている\cite{Moerland2018-kk}。
%それぞれの論文を参考に、関連論文の扱う感情カテゴリについて、Ekmanの基本６感情（anger, disgust, fear, happiness, sadness, surprise, neurtal）以外のものを表\ref{table:category}にまとめる。
%表と社会的感情の定義より、示されている感情を照らし合わせると、shameやpride, admirationが存在することがわかるが、これらの研究の中にはその他の社会的感情はみられなかった。
%しかし、今回は社会的感情と基本的感情の扱いの分類が困難であるAngerなどは除いていることに注意が必要である。
%101件の論文が対象となった を追加
In this section, emotions research in robotics are briefly summarized. 
Yan {\it et al.} investigated the emotional space of social robots \cite{Yan2021-zf}. 
Cavallo {\it et al.} reviewed emotions in social robots and revealed that many studies on social robots are based on Ekman's basic emotions \cite{cavallo2018emotion}. 
Moerland {\it et al.} summarized emotion model research using robots and agents targeting reinforcement learning \cite{Moerland2018-kk}. 
With reference to the aforementioned literature (101 articles were targeted in three survey papers) , Table \ref{table:category} summarizes the emotion categories that have been addressed by the related studies, except for Ekman's six basic emotions (anger, disgust, fear, happiness, sadness, surprise, and neutral). 
The emotion categories shown in these studies (and listed in the table), as well as the definitions of social emotions, reveal the presence of shame, pride, and admiration; however, no other social emotions were present. 
Note that anger, which is difficult to distinguish between a social and a basic emotion, was excluded. 

\begin{table}[t]
 \caption{Emotion category in robots and agents research (excluding basic emotions)}
  \label{table:category}
  \centering
  \begin{tabular}{|l|c|} \hline
    Emotion & Research  \\ \hline \hline
    Joy &  \cite{jacobs2014emergent}, \cite{shi2012artificial}, \cite{el2000flame}, \cite{moussa2013toward} \\ \hline
    Hope &  \cite{jacobs2014emergent}, \cite{el2000flame}, \cite{moerland2016fear}, \cite{lahnstein2005emotive}\\ \hline
    Frustration & \cite{hasson2011emotions}, \cite{huang2012goal}, \cite{tsankova2002emotionally} \\ \hline
    Distress & \cite{jacobs2014emergent}, \cite{moussa2013toward}\\ \hline
    Relief &  \cite{shi2012artificial}, \cite{el2000flame} \\ \hline
    Gratitude, Reproach, Admiration, Pride, & \cite{el2000flame} \\ 
    Shame, Gratification, Remorse &   \cite{moussa2013toward}\\ \hline
    Boredom & \cite{goerke2006emobot} \\ \hline
    Comfort & \cite{blanchard2005imprinting} \\ \hline
    Contented, Elation, Panic &  \cite{doshi2004towards}\\ \hline
    Cooperative, Slightly, Annoyed &  \cite{gmytrasiewicz2002emotions} \\ \hline
    Disappointment & \cite{el2000flame} \\ \hline
    Resentment, Sorry for, Gloating & \cite{moussa2013toward} \\ \hline
  \end{tabular}
\end{table}

In addition, with respect to the dimension theory of emotions, several studies involving robots have used VA, which were famous in Russel's circumplex model. 
It cannot be denied that social emotions may be included in VA. However, VA cannot distinguish between social and basic emotions. In addition, there have been studies which used PAD \cite{mehrabian1974approach,lim2014mei,bera2019emotionally,fang2018personality,claret2017exploiting} and VAS \cite{breazeal1999build,jing2015cognitive,lun2016cognitive,liu2017empathizing}. 
%ここが英語難しい。。。
%Although they considered others, they showed the directionality of relationships with others and things, and were not themselves the axes that represent oneself and others.
Although social ``others'' were involved in these research projects, the existence of others represented the directionality of relationships between the robot and others. 
There was no axis that represented self/other in the emotional space, as has been proposed in social emotion research. 

There have been few robot studies that are directly based on social emotions. 
For example, Tsiourti {\it et al.} conducted a comparative study to investigate how humans perceive emotional cues expressed by humanoid robots through five communication modalities (face, head, body, voice, and locomotion) \cite{tsiourti2017designing}. 
This study addressed three emotions: happiness, sadness, and surprise. The authors argued that these three emotions are social emotions according to the definition of Harelli {\it et al.} \cite{ hareli2008s}. 

\subsection{Survey based on categories of the social emotion}
We also conducted a mechanical survey of articles on robots using keywords regarding the social emotions (embarrassment, shame, guilt, jealousy/envy, love, fago, grief, pride, pridefulness, admiration, liking, and contempt) that were described in Section \ref{sec_def}. 
The target articles are those of the 2000s. 
%2000年代の論文をターゲットとした。（これより古いのは引いていない）
Consequently, shame, jealousy/envy, love, grief, pride, admiration, and contempt were found in the emotion recognition and expression of robots, as summarized in Table \ref{table:robot}. 
Regarding embarrassment and guilt, psychological experiments have been conducted in which specific scenarios were presented to humans to evaluate whether humans have those feelings toward robots \cite{bartneck2010influence,aymerich2019your}. 
However, we could not identify any research wherein the existence of that emotion was actually incorporated as part of the robot system. 

%%%%%%%
\begin{table}[t]
 \caption{Social emotions and robots}
  \label{table:robot}
  \centering
  \begin{tabular}{|l|c|} \hline
    Emotion & Research  \\ \hline \hline
    Shame & EL-Nasr et al., 2000 \cite{el2000flame}; Moussa et al., 2013 \cite{moussa2013toward};  \\ 
     & Hegel et al., 2010 \cite{hegel2010social} \\ \hline
    Jealousy / Envy & Oliveira et al., 2019 \cite{oliveira2019stereotype} \\ \hline
    Love & Samani, 2016 \cite{samani2016evaluation} \\ \hline
    Grief & Huahu et al., 2010 \cite{Huahu2010App}; Terada et al., 2012 \cite{terada2012artificial} \\ \hline
    Pride & EL-Nasr et al., 2000 \cite{el2000flame}; Moussa et al., 2010 \cite{moussa2013toward};  \\ 
     &  Oliveira et al., 2019 \cite{oliveira2019stereotype} \\ \hline
    Admiration & EL-Nasr et al., 2000 \cite{el2000flame}; Moussa et al., 2010 \cite{moussa2013toward};  \\ 
     &  Oliveira et al., 2019 \cite{oliveira2019stereotype}; Terada et al., 2012 \cite{terada2012artificial} \\ \hline
    Contempt & Oliveira et al., 2019 \cite{oliveira2019stereotype}; Jung, 2017 \cite{jung2017affective} \\ \hline
  \end{tabular}
\end{table}
%%%%%%%

From Table \ref{table:robot}, El-Nasr {\it et al.} addressed pride, shame, and admiration in robots \cite{el2000flame}. 
In that study, pride was defined as an action done by the agent that was approved by the available standards, shame was an action done by the agent that was disapproved by the standards, and admiration was an action done by another agent that was approved by the agents' standards. 
In \cite{moussa2013toward}, pride, shame, and admiration were used in the system to determine an agent's facial expression from human facial expressions and audio information. 
In addition, Hegel {\it et al.} had robots express shame via LEDs on the robot's cheeks as the robot that performed different facial expressions \cite{hegel2010social}. 

Oliverira {\it et al.} presented groups of two people and two robots with a card game to play. 
For the robots, two characteristics of competence and warmth were divided into two levels, high and low. The researchers designed behavior for the robots corresponding to each characteristic, and the reaction of human beings to the behavior of the robot was investigated. 
Here, jealousy/envy was expressed as high competence and low warmth, pride and admiration were expressed as high competence and high warmth, and contempt was expressed as low competence and low warmth \cite{oliveira2019stereotype}. 

Regarding love, previous researchers have attempted to validate user verification of love between humans and robots with reference to human-human affection \cite{samani2016evaluation}. Here, the robot received audio, visual, tactile, and acceleration information and interacted with a person using motion, audio, and LED. 
In addition, in the research by Huahu {\it et al.}, robots recognized grief as emotional recognition from human utterances \cite{Huahu2010App}.
Jung explained that ``turning against'' is recognized as contempt from an affective standpoint in the interactions between robots and humans \cite{jung2017affective}. 

Terada {\it et al.} used a glowing robot to express grief and admiration \cite{terada2012artificial}.
In the study by Terada {\it et al.}, experiments were based on Plutchik's wheel of emotions. 
In Plutchik's Wheel of Emotions, grief and admiration are listed as part of the basic emotions \cite{athar2011fuzzy} (Table \ref{table:plutchik}). 
In Plutchik's model, there are other applied emotions, such as shame, guilt, envy, love, pride, and contempt (Table \ref{table:plutchik_high}). 
These are composed of a combination of basic emotions.
Plutchik's model is similar to dimensional theories, such as the VA mentioned above, and in this model social emotions are not separated from basic emotions. 

\begin{table}[t]
 \caption{Basic emotions of Plutchik's Wheel of Emotions}
  \label{table:plutchik}
  \centering
  \begin{tabular}{|l|l|} \hline
    \bf{Emotion} & \bf{Content}  \\ \hline \hline
    Joy & Basic emotion \\ \hline
    Trust & Basic emotion \\ \hline
    Fear & Basic emotion \\ \hline
    Surprise & Basic emotion \\ \hline
    Sadness & Basic emotion \\ \hline
    Disgust & Basic emotion \\ \hline
    Anger & Basic emotion \\ \hline
    Anticipation & Basic emotion \\ \hline \hline
    Ecstasy & Strong joy \\ \hline
    \bf{Admiration} & Strong trust \\ \hline
    Terror & Strong fear \\ \hline
    Amazement & Strong surprise \\ \hline
    \bf{Grief} & Strong sadness \\ \hline
    Loathing & Strong disgust \\ \hline
    Rage & Strong anger \\ \hline
    Vigilance & Strong anticipation \\ \hline \hline
    Serenity & Week joy \\ \hline
    Acceptance & Week trust \\ \hline
    Apprehension & Week fear \\ \hline
    Distraction & Week surprise \\ \hline
    Pensiveness & Week sadness \\ \hline
    Boredom & Week disgust \\ \hline
    Annoyance & Week anger \\ \hline
    Interest & Week anticipation \\ \hline
  \end{tabular}
\end{table}    

\begin{table}[t]
 \caption{Higher-level emotions of Plutchik's Wheel of Emotions}
  \label{table:plutchik_high}
  \centering
  \begin{tabular}{|l|l|} \hline
    \bf{Emotion} & \bf{Blend}  \\ \hline \hline
    \bf{Love} & Joy + Trust \\ \hline
    Submission & Trust + Fear \\ \hline
    Alarm / Awe & Fear + Surprise \\ \hline
    Disappointment & Surprise + Sadness \\ \hline
    Remorse & Sadness + Disgust \\ \hline
    \bf{Contempt} & Disgust + Anger \\ \hline
    Aggressiveness & Anger + Anticipation \\ \hline
    Optimism & Anticipation + Joy \\ \hline \hline
    \bf{Guilt} & Joy + Fear \\ \hline
    Curiosity & Trust + Surprise \\ \hline
    Despair & Fear + Sadness \\ \hline
    Unbelief & Surprise + Disgust \\ \hline
    \bf{Envy} & Sadness + Anger \\ \hline
    Cynicism & Disgust + Anticipation \\ \hline
    \bf{Pride} & Anger + Joy \\ \hline
    Hope & Anticipation + Trust \\ \hline \hline
    Delight & Joy + Surprise \\ \hline
    Sentimentality & Trust + Sadness \\ \hline
    \bf{Shame} & Fear + Disgust \\ \hline
    Outrage & Surprise + Anger \\ \hline
    Pessimism & Sadness + Anticipation \\ \hline
    Morbidness & Disgust + Joy \\ \hline
    Dominance & Anger + Trust \\ \hline
    Anxiety & Anticipation + Fear \\ \hline
  \end{tabular}
\end{table}

\subsection{Mapping between existing robots and social emotions}
\begin{figure}[t]
  \begin{center}
    \includegraphics[width=12.0cm]{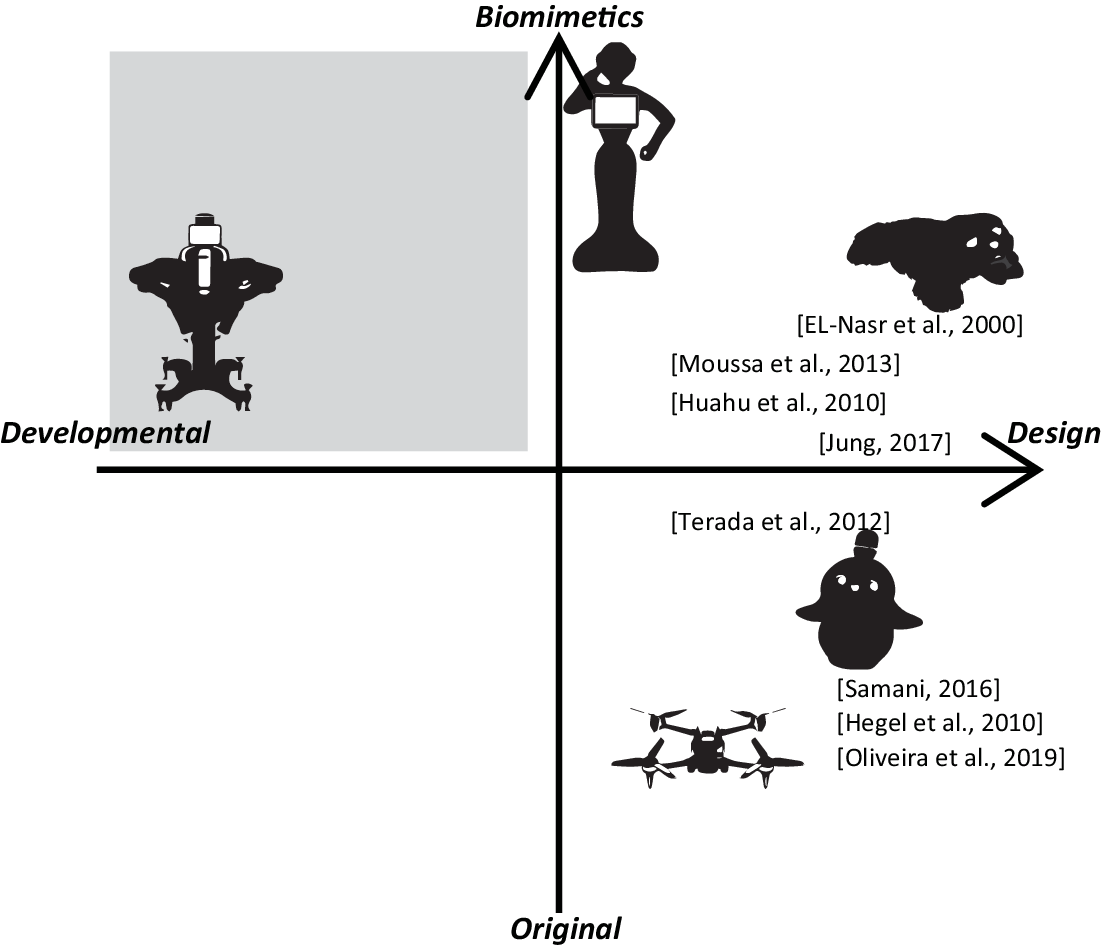}
    \caption{Approach to robot's social emotions}
    \label{fig:outro}
  \end{center}
\end{figure}

%図の追加
%導入で示した図\ref{fig:intro}の空間に調査した表\ref{table:robot}の研究を当てはめると図\ref{fig:outro}のようになる。図\ref{fig:intro}で示した研究は社会的感情を視野に入れているものはあったものの、明に社会的感情を実装してはおらず、図\ref{fig:outro}では記載をしていない。
%この結果をみると、導入で大事だと主張した第二象限の研究がないことがわかる。感情モデル研究は多く行われているが、明に社会的感情を実装している研究はみつけられなかったことを表す。
%しかし、そもそも社会的感情を扱う研究自体が少ないため、まずはデザインで増えていき、これから発展していくとも考えられる。
Figure \ref{fig:outro} was obtained by applying the research presented in Table \ref{table:robot} to the space shown in Fig. \ref{fig:intro} in the introduction. 
Although some of the studies shown in Fig. \ref{fig:intro} have taken social emotions into consideration, they do not clearly implement social emotions and thus are not shown in Fig. \ref{fig:outro}.
Looking at this result, we can see that there has been no study in the second quadrant, which we identified in the introduction as being particularly important. 
Although many emotion modeling studies have been conducted, we were unable to find any studies which clearly implemented social emotions.
However, because there are a few studies which have addressed the topic of social emotions, we believe that the number of studies investigating social emotions in robots will increase as a result.

%%%%%%%
%%%%%%%

\section{Discussion}
Based on our observations, there have been few robot studies on social emotions. 
Even in the studies which have focused on social emotions, they did not differentiate basic emotions from social emotions. 
Instead, they treated the two types of emotions as conforming to the same dimension, and designed expressions and recognition procedures based on these rules. 
Thus, present social robot research conflicts with the results from psychological-neurological findings. 
We believe this is not necessarily an incorrect wrong direction of research, considering that robots address social emotions superficially. 
A simple system can be constructed by assuming that all emotions are in the same layer without separating complex emotions from the more basic emotions, and if the goal is to simply convey the emotional expression to others, this purpose may be achieved even with a simple rule-based implementation. 
However, it is difficult to determine whether all complex, higher-level emotions can be realized by manually designing their rules. 
It is difficult to imagine that such an approach would be able to replicate sufficiently complex emotional behavior. 
Moreover, as mentioned above, there is a gap between the current psychological-neurological findings and the implementation of social emotions in current robots. 
To achieve emotional development, we must fill that gap.
Here, we discuss future challenges based on surveys from this perspective. 

\subsection{Current status of emotions research using robots}
%\ref{sec:pre_robot}では、既存のロボット研究について述べた。
%その後の社会的感情を扱うロボット研究の調査結果では、ここで挙げたほとんどの研究が明に社会的感情を実装していないことが明らかとなった。
\ref{sec:pre_robot} summarized existing emotion studies using robots. 
Subsequent survey of research on robots dealing with social emotions revealed that most of the studies listed here clearly did not implement social emotions. 
%特に重要視していた第二象限が社会的感情を扱うものをみつけられなかった。
%デザインによって実装したものは存在するもののその数は少なく、まだほとんど社会的感情を扱うロボット研究がなされていないことを示している。
We could not find any research that dealt with social emotions in the second quadrant of our figure, which is of particular importance.
This indicates that, although there have been some successful design implementations, the number of such projects is small and robot research dealing with social emotions has not been conducted. 
%たとえば、pepperは感情地図によって、大きな視野をもって研究を進めているが、実際に実装されているのは基本感情に留まっている。
%これはデザインによる限界を表しているとも考えられる。
For example, in pepper, research is being carried out from a broad perspective using emotion maps, but only basic emotions are being implemented. 
This may represent a limitation of implementation by design.

%また研究とは言えないが、lovotは嫉妬を実装している。
%これに関しては完全なデザインであり、実際にこれが他者に嫉妬と捉えられるかなどの検証もしていないため、社会的感情を扱う研究として挙げることはできない。
``Jealousy'' has been implemented in lovot. 
This is a complete design, however it has not been verified as to whether it is actually perceived as jealousy by humans, so it cannot be cited as a valid example of research dealing with social emotions. 
%さらに、このような感情のデザインによる実装は嫉妬の本質を捉えられているとは思えないし、異なるシチュエーションに関して、lovotは嫉妬を表出せず、自身が嫉妬しているという事実も厳密には認識していない。また、この嫉妬の行動によってロボットが利を得ることもないのだろう。
It is unlikely that such an emotional design implementation is able to capture the essence of jealousy. 
In different situations, lovot does not express jealousy and is unaware of the fact that lovot is itself jealous. 
Furthermore, this jealousy behavior does not benefit the robot itself. 
%感情は本来は生存システムに組み込まれていると考えられ、人はある種の報酬を得るために感情を発達させていると考えられる。
%これに根差さないシステムは、感情の本来の役目を果たしているとはいえないだろう。
%そういう意味で、Fig. \ref{fig:intro}で示した第二象限の研究群を発展させることは、社会的感情を明に示す可能性のある手法だと考えられる。
Emotions are thought to be inherently built into the survival system, and one is thought to develop emotions as a means toward some reward. 
A system that is not rooted in this survival system cannot be said to play the original role of emotions.
In that sense, developing the research in the second quadrant shown in Fig. \ref{fig:intro} would be a useful method that may clarify the essence of social emotions. 
%%%%%%%%
\subsection{Can we still design social emotions in robots?}
%残念ながら，現状では感情の発達的なモデルを実装することでロボットに社会的感情を持たせる道筋は容易なものではない．
%したがって，パートナーロボットを作るために，デザインによって社会的感情を作りこむことは，状況を限定し複雑さを犠牲にすれば近道であるともいえる．
%実際，上述のlovotの例はそれを具現化しているが，その成否は現状では明らかでない．
Unfortunately, at present, it is not easy to provide robots with social emotions by implementing a developmental model of emotions.
Therefore, creating social emotions by design to build a partner robot is a reasonable method, as long as we acknowledge that it will applicable in limited situations and with simple behaviors. 
In fact, the above example of lovot is a realization of this style of approach, but its success or failure is not clear at present. 

%感情カテゴリの正解についての意見
%デザインアプローチを取る際に，どのような感情カテゴリを採用すべきであろうか？
%本論文では感情状態に関して様々なものを挙げたが、これらのどれが正解かは示すことができない．
%なぜなら，本質的に感情のカテゴリはインタラクションの中で創発していくと考えられ，本論文の立場では，強固な正解は存在しないことになる．
%我々は，感情の創発という立場で外受容感覚や内受容感覚のような多次元の空間が言語コミュニケーションによって概念化されるような空間を感情状態の空間と考えている．
%感情の基盤は，内受容感覚のある種の概念である．
Which emotional categories should be adopted when taking a design approach? 
In this paper, we have listed various emotion categories, however, we cannot determine which of these is the correct answer. 
This is because, in essence, the emotional category is considered to emerge through interactions, and from the standpoint of this paper, there is no ground truth.  
From the viewpoint of emotion emergence, we consider a space in which multidimensional information, such as extraceptions and/or interoceptions, is conceptualized through verbal communication as an emotional space. 
The basis of emotions is a kind of concept of interoceptions.

%
%こうした点からも，言語カテゴリだけで感情を記述することには問題があるかもしれないという社会構築主義における指摘は重要である \cite{Aranguren2016}．
%しかし，従来示されてきたカテゴリが正しくないということを主張しているわけではない．
%次元説についても，いずれかの側面で次元圧縮した結果であり，本質的にはより多次元の空間であると考えられる．
%従って，ロボットに社会的感情をデザインするという立場をとる場合，どの感情カテゴリを使うべきかは，アプリケーションに依存し選択することになる．
From this point of view, it is important to note that there is a problem in describing emotions only by language category, which has been previously identified by social constructivists \cite{Aranguren2016}. 
However, we do not claim that the traditionally presented categories are incorrect.
The dimensional theory is also the result of dimensionality reduction in one aspect, and it is thought that emotions are essentially expressed in a higher dimensional space.
Therefore, when taking the approach of designing social emotions for a robot, which emotion category should be used depends on the application.
%社会構成主義的な主張は，ロボットを用いた感情研究に対しても当てはまる．
%そもそも，ロボットに感情をデザインすることは，感情の社会的な機能を実現しようとしてると捉えることができる．
%これはまさに社会構成主義的な発想であり，そうしたデザインは社会構成主義を突き詰めていく必要がある．
%現状のロボットにおける感情は，ある種の``Emotionology Principle''に基づいており，より汎用的な感情表現をデザインするためには、オントロジーとメソドロジーの分離が必要であると思われる．

Now, it is important to be aware that the social constructivist claims can directly apply to emotional design approaches when using robots. 
Designing emotions on robots can be regarded as attempting to realize the social functions of emotions.
This is exactly the idea of the social constructivist, and such a design must pursue the social constructivist view of emotions. 
Considering the descriptions in \ref{sec:social}, we can see that emotions in current robots are based on a kind of ``Emotionology Principle.'' 
%この``Emotionology Principle''の考え方は，自然言語処理に例えると理解しやすい．
%自然言語処理では，言葉の意味を関係性の中に見出そうとする．
%例えば，Word2Vecの技術であれば，各単語が文章の中でどのように使われるかを規範に，単語同士の意味的な距離をニューラルネットワークが学習する \cite{mikolov2013efficient}．
%Emotionology Principleも同様に，感情を表す語が，どのような状況で使われるかを詳細に調べることで，その本質が理解できると考える．
%この，``状況''をくまなく書き下し、各状況における反応を定義すれば社会的感情はデザインできることになるが，これは非常に困難な作業であり人手では不可能である．
The idea of the emotionology principle is that it is easy to understand when compared to natural language processing (NLP). 
NLP basically attempts to find the meaning of words in relationships. 
For example, in the case of Word2Vec technology, a neural network learns the semantic distance between words based on how each word is used in sentences \cite{mikolov2013efficient}. 
Similarly, the emotionology principle argues that the essence of emotions can be understood by examining in detail the situations in which emotional words are used. 
Social emotions can be designed by writing down all these situations and defining the reaction in each situation. 
However, this is a very difficult task which cannot be done manually. 

%一方で現在のNLPの大きな成功は，機械学習の技術と大量の言語コーパスに支えられている．
%従来不可能であると考えられていた雑談対話も，かなりの水準で実現されつつある \cite{xu2021goldfish}．
%同様のことがロボットへの感情の実装の場面で応用可能であろうか？
%問題として，大きく二つの点が挙げられる．
%一つは，大量のデータを集められるのかという問題である．
%例えばこれは，人の生体データとその時の行動，および状況を表現するための一人称視点映像を大量に収集することを意味する．
%二つ目は，個人性の問題である．
%こうして考えると，機械学習を援用した社会的感情のデザインも，可能性は感じるものの容易な方向性であるとは言い難い．
Alternatively, the great success of NLP today is supported by machine learning technology and huge language resources.
Chatbot systems, which were previously thought to be impossible, are being developed to considerable performance levels \cite{xu2021goldfish}. 
Is the same applicable in the context of implementing emotions in robots?
Here, there are two main problems.
First is the question of whether a large amount of relevant emotion data can be collected.
For example, this would require collecting a large amount of human physiological data and first-person view images to identify the agent's behavior and their situation at the time that they experience different emotions.
The second is the issue of individuality.
Considering these problems, it is difficult to determine if the design of social emotions using machine learning is an easy direction for future research, although it at least seems like a possible approach.

%いずれにしても，各個体のもつ社会的感情のメカニズムに迫るためには，構成的アプローチが重要である．
%以降は，そうした視点で議論する．
In any case, the constructive approach is more important than the design approach for revealing the mechanisms of social emotions of individuals.
For the rest of this paper, we will discuss issues from that perspective. 
%and it seems necessary to separate ontology and methodology to design more general-purpose emotional expressions. 
%emotionologyの考え方

\subsection{What is missing in current studies on social emotions in robotics?}
First, let us summarize the difference between psychological and neurological findings, and the implementation of social emotions in current robots. 
From the definition of social emotions, Elster classified social emotions based on the criteria of self-other evaluations, as shown in Table \ref{table:elster}. 
In the self-conscious emotion, which is a subcategory of social emotions, it is stated that self-awareness is the awareness of being different from others, that is, self-other discrimination, is significant.
Thus, regarding social emotions, the existence of others is considered an important factor in creating social emotions.
However, the papers listed in Table \ref{table:robot}, which currently deal with social emotions, do not include the elements of others in their definitions of emotions. 
These studies assume a context of human-robot interaction (HRI), at which point each robot behaves in the social context of the interaction. 

From a perspective based on psychological findings, it has been suggested that factors such as others, groups, and evaluations through others, influence the emergence of social emotions. 
For example, in the ``It--Me'' axis shown in Figure \ref{fig:itme}, social emotions are categorized according to different factors, such as whether the emotion is directed toward oneself or the other. 
According to neurological findings, the insula (which acts as a hub in emotional activities), PSTS, TPJ, and DMPFC (which is involved in mentalizing and acquiring the perspective of others) are involved in interpersonal and social emotions. 
According to Table \ref{table:part}, the active brain regions are different between individual emotions and social-interpersonal emotions, and it can be observed that common but different activities occur in the brain for these different emotion types. 
However, the robotics studies surveyed in this paper do not mention these important points, such as acquiring the perspective of others. 
Because these studies mainly focused on HRI, it is believed that they are addressing interpersonal emotions, which are a subset of social emotions. 

\subsection{Challenges toward the implementation of social emotions}
\label{subsec:selfother}
What are the necessary elements to implement social emotions in robots? 
From the current discussions, we believe that elements of self-other discrimination, the acquisition of the viewpoint of others, and mentalizing are necessary. 
Thus, the mirror neuron system \cite{RizzolattiBook2008} plays an important role. 
These elements have been studied as constructive approaches in cognitive developmental robotics \cite{nagaiyukie2011,Gordon2020}. 
However, current research has not progressed to studying social emotions.

%つながりは後で調整
%重要な点は，社会的感情は人の社会性の根源である自己や他者の同一視からの分離，つまりはミラーニューロンシステムに基づく他者の理解と、他者視点の獲得に深く関わっていると考えられることである．
%従って，社会的感情をもつ，もしくは発達させるために，人間が備える機能をロボットで実現可能かという点は重要なポイントである．
%例えば，ミラーニューロンシステムを最も単純に工学的に実現するとすれば，OpenPoseに代表される人のポーズ推定技術 \cite{openpose,kudo2018} が利用可能であると考えられる．
%そうしたアルゴリズムを使うことで，他者の運動を自身の関節の運動に投影して認識することが可能となる．
%さらに，Generative Query Network (GQN) \cite{EslamiGQN2018} は，3次元世界の様々な位置での視点を生成することのできる技術であり，例えば他者の位置情報から，その位置における世界の見え方を推定することを原理的に可能としている点で，他者視点の獲得を工学的に実現できる可能性を示している．
%こうした，近年の深層学習による技術の進歩は，社会的感情を本質的に作るという方向性において非常に重要な要素となると考えられる．
The important point is that social emotions are deeply involved in the separation of self and others from their identification, which is the root of human sociality. That is, the understanding of others based on the mirror neuron system and the acquisition of the perspective of others.
Therefore, it is an important point whether robots can realize these functions that humans have, in order to have or develop social emotions. 
For example, the simplest engineering realization of a mirror neuron system would be to use human pose estimation techniques such as OpenPose \cite{openpose,kudo2018}. 
By using such an algorithm, it is possible to project and recognize the movements of others on the movements of one's own joints. 

Furthermore, the Generative Query Network (GQN) \cite{EslamiGQN2018} is a technology that can generate viewpoint images at various positions in the three-dimensional world. 
For example, using GQN it is possible to estimate the appearance of the world at a particular position from the position information of others. 
Thus, in principle, it is possible, from an engineering perspective, to acquire the viewpoint of others.
These recent technological advances through deep learning are very important factors in the direction of creating social emotions. 
%
%acquisition of the viewpoint of others
%Regarding the acquisition of the viewpoint of others, the neural scene representation, which made great progress recently, called Generative Query Network (GQN) \cite{EslamiGQN2018} is a promising key technology to its realization. 
In the future, it will be necessary to study, through robots, mechanisms that develop into social emotions by linking self-other discrimination, acquisition of the perspective of others, and mentalizing centered on the mirror neuron system with emotions, which all involve conceptualizations of the physical body.

Additionally, the study of empathy in robots is important. 
For example, Asada proposed a developmental model of artificial empathy \cite{asada2015towards} and described the development of self-other cognition. 
First, self-awareness occurs by manipulating objects. After synchronization with other persons using mirror neurons, it is observed that the other person behaves differently, and self-other discrimination occurs. 
Further, Asada explained that this factor could generate emotions, such as envy and schadenfreude. 
Empathy would therefore be considered as a type of synchronization mechanism, although it is believed that various emotions are generated when changing from synchronization to desynchronization.

\subsection{Challenges toward emotion development in robots}
The authors constructed an emotion model to create social emotions with robots \cite{hieida2018deep}. 
In this study, we argued that when an action is output, an emotional category is formed by categorical input (for example, language) from another person. 
We simulated emotional differentiation using a task that mimics the interaction between a caregiver and a child. 
In that study, only basic emotions were addressed; however, according to the idea of Lewis mentioned above, after the emergence of basic emotions, self-other discrimination is performed, and then social emotions can emerge. 
Therefore, we believe that social emotions can be achieved by realizing the separation of self and others with this model. 

Simply put, there is a possibility that social emotions can emerge by integrating the self-other separation algorithm, as described in Section \ref{subsec:selfother}, for example,\cite{Gordon2020}, with the authors' algorithm. 
Whether it has human-like basic emotions and social emotions depends on the body and environment of the robot in which it is implemented.
The design of the robot body, including soft robots for human-like emotions, is an important research direction. 

Furthermore, this is a problem related to the intelligence of robots that integrate various senses and decisions, not just emotions. 
Thus, we believe that it is necessary to study which types of bodies, environments, interactions, and algorithms produce which types of intelligence with an actual robot. 
We believe it is necessary to connect the robot technology realized in this manner to the moral problems, which will be discussed in \ref{sec:moral}.

\subsection{Social constructivist view of emotions for the developmental approach}
%発達アプローチにおける社会的構成主義的視点
%\ref{sec:social}で述べたように，感情は社会的なインタラクションや文化に大きな影響を受けて形成される．
%特に社会的な感情において，社会や文化は大きな影響を持つと思われる．
%従って，\ref{subsec:selfother}で述べた技術的要素に，こうした社会的影響を取り込むことは，社会的感情の研究において必要不可欠である．
As mentioned in \ref{sec:social}, emotions are formed under the influence of social interactions and culture. 
Society and culture seem to have a great influence, especially on social emotions. 
Therefore, incorporating such social influences into the technical elements mentioned in \ref{subsec:selfother} is indispensable for the study of social emotions.

%しかし，社会や文化との関係を構成的に再現することは，非常に難しい課題である．
%これは，言語の学習とも関連しており，知能全体としての構築が必要になると考えられる．
%そのためにも，記号創発ロボティクスのアプローチが参考になる\cite{SER2016,SER2019}．
%ただし，記号創発ロボティクスにおいても，社会や文化との関りは扱い始めたばかりの課題であり，感情の問題を取り込むためにはまだまだ多くの理論的整備やロボットの性能向上といった課題を解決する必要がある．
%
%さらに重要なことは，こうして工学的に実現した感情と，人間を対象にして調べた感情の比較によって，新たな知見や問題を明らかにすることである．
%そうすることで，構成的アプローチの本来の役割を果たす必要がある．
However, it is very difficult to constructively reproduce the relationship with society and culture in an experimental setting. 
This is also related to language learning, and it is thought that it is necessary to build intelligence as a whole.
For that purpose, the approach of symbol emergence in robotics (SER) is of great help \cite{SER2016,SER2019}. 
Even in SER, the relationship between society and culture is a problem that has just begun to be addressed. Therefore, to incorporate emotional problems, it is necessary to solve many of the existing problems, including the improvement of robot performance. 
More importantly, it is necessary to clarify new findings and problems by comparing the emotions realized by engineering with the emotions examined in humans.
In doing so, the constructive approach will play an essential role. 

\subsection{Moral emotions in robotics}
\label{sec:moral}
In the section on the definition of social emotions, moral emotions were introduced as a subcategory of social emotions; however, they is not covered in this study. 
Moral emotion is an interesting topic in view of recent attention to morality in HRI \cite{abuse2015,nomuraAbuse2016,BartneckKeijsers+2020+271+283}. 
Moral emotions are triggered by the perception of moral violations in the context of interpersonal events, and they can guide moral behavior. Moral emotion is an important factor in encouraging the moral behavior of the robot. This category includes shame, guilt, regret, embarrassment, contempt, anger, disgust, gratitude, envy, jealousy, schadenfreude, admiration, sympathy, and empathy. Empathy, which is treated as an element that creates social emotions, is treated as a part of moral emotions. Moreover, other complicated emotions, such as schadenfreude, are also included in moral emotions. However, how to address these differences in each definition remains controversial. 

The following can be considered as issues in future moral emotion research in robotics: 1) Is it possible to make a robot moral, and for robots to understand morals, by implementing moral emotions? 2) Can we better understand the mechanism of moral emotion through robots? 3) Can we understand how to nurture people's morals and make people behave more morally?
Considering the coexistence of humans and robots, it is necessary to discuss both sides of the morals of humans and robots in an integrated manner. 

\section{Conclusion}
%本論文では社会的感情とロボットに関する研究を調査し報告した。感情全般および社会的感情の定義から始まり、心理学的知見ならびに神経科学的知見、ロボティクスでの社会的感情の扱われ方、それを踏まえた上でロボティクスと他の分野の知見との乖離、ロボットに社会的感情を実装する上で必要な要素について議論した。本分野はまだまだブルーオーシャンである。如何に基本感情との立ち位置を考えつつ、実装していくかは大きな課題であり、このことが感情発達メカニズムの解明の一助となると考えている。
%コメントで研究の応用範囲や将来展望をもっとかけとの指摘があったので何書こう…
This survey paper examined research on social emotions in the area of robotics. 
We discussed the definitions of social emotions, psychological-neurological findings, how social emotions are treated in robotics, the gap between robotics and knowledge in other research fields, and challenges for implementing social emotions in robots. 
This research field is still developing. 
A major issue is how to implement social emotions in robots while considering their relationship with basic emotions. 
We believe this will help elucidate the mechanisms underlying emotional development.
Furthermore, we hope to better understand the morals of humans and robots through social emotions, and to explore ways which lead to a better future in which humans and robots can coexist harmoniously.

\section*{Acknowledgement(s)}
This work was supported by JSPS KAKENHI Grant-in-Aid for Early-Career Scientists (20K19907) and JSPS KAKENHI Grant-in-Aid for Scientific Research on Innovative Areas (20H05565).

\section*{Notes}
This is a preprint of an article submitted for consideration in ADVANCED ROBOTICS, copyright Taylor \& Francis and Robotics Society of Japan; ADVANCED ROBOTICS is available online at http://www.tandfonline.com/.

\bibliographystyle{tfnlm.bst}
\bibliography{ARhieida2021.bib}

\end{document}